\documentclass{article}

\usepackage[final]{corl_2020} 
\usepackage{xspace}
\usepackage{microtype}
\usepackage{graphicx}
\usepackage{mathtools}
\usepackage{siunitx}
\usepackage[font=small]{caption}
\usepackage{wrapfig}
\usepackage{booktabs} 
\usepackage{amssymb}
\usepackage{amsmath,amsthm}
\usepackage{mathtools}
\usepackage{subcaption}
\usepackage{algorithm}
\usepackage{algpseudocode}
\usepackage[utf8]{inputenc}
\usepackage[english]{babel}
\usepackage{enumitem}
\newtheorem{theorem}{Theorem}[section]
\newtheorem{lemma}{Lemma}[section]

\theoremstyle{definition}
\newtheorem{definition}{Definition}[section]
\newtheorem{assumption}{Assumption}[section]
\usepackage{titlesec}
\usepackage[resetlabels]{multibib}
\newcites{SM}{References}
\newcites{covid}{References}

\titlespacing\section{1pt}{1pt plus 2pt minus 2pt}{1pt plus 2pt minus 2pt}
\titlespacing\subsection{0pt}{3pt plus 4pt minus 2pt}{0pt plus 2pt minus 2pt}
\titlespacing\subsubsection{0pt}{3pt plus 4pt minus 2pt}{0pt plus 2pt minus 2pt}
\usepackage{autolabtools}

\usepackage{cleveref}

\DeclareMathOperator*{\argmax}{argmax}

\newcommand{\Ex}{\mathbb{E}}

\title{\probname:\\ Asymptotically Optimal Algorithms for \\ Grasping Challenging Polyhedral Objects}

\author{
  Michael Danielczuk*, Ashwin Balakrishna*, \\\textbf{Daniel S. Brown, Shivin Devgon, Ken Goldberg}\thanks{The AUTOLab at UC Berkeley (\url{automation@berkeley.edu})}
  \\\scriptsize{* equal contribution} \\
  \texttt{\{mdanielczuk, ashwin\_balakrishna\}@berkeley.edu}\\
}
\makeatletter
\def\thanks#1{\protected@xdef\@thanks{\@thanks
        \protect\footnotetext{#1}}}
\makeatother
\begin{document}
\newcommand{\probname}{Exploratory Grasping\xspace}
\newcommand{\algname}{Bandits for Online Rapid Grasp Exploration Strategy (BORGES)\xspace}
\newcommand{\algabbr}{BORGES\xspace}
\newcommand\smallO{
  \mathchoice
    {{\scriptstyle\mathcal{O}}}
    {{\scriptstyle\mathcal{O}}}
    {{\scriptscriptstyle\mathcal{O}}}
    {\scalebox{.6}{$\scriptscriptstyle\mathcal{O}$}}
  }

\newcommand{\ashwin}[1]{\textcolor{magenta}{(#1 --Ashwin)}}
\newcommand{\mike}[1]{\textcolor{blue}{(#1 --Mike)}}
\newcommand{\daniel}[1]{\textcolor{orange}{(#1 --Daniel)}}
\maketitle


\begin{abstract}
There has been significant recent work on data-driven algorithms for learning general-purpose grasping policies. However, these policies can consistently fail to grasp challenging objects which are significantly out of the distribution of objects in the training data or which have very few high quality grasps. Motivated by such objects, we propose a novel problem setting, \probname, for efficiently discovering reliable grasps on an unknown polyhedral object via sequential grasping, releasing, and toppling. We formalize \probname as a Markov Decision Process where we assume that the robot can (1) distinguish stable poses of a polyhedral object of unknown geometry, (2) generate grasp candidates on these poses and execute them, (3) determine whether each grasp is successful, and (4) release the object into a random new pose after a grasp success or topple the object after a grasp failure. We study the theoretical complexity of \probname in the context of reinforcement learning and present an efficient bandit-style algorithm, \algname, which leverages the structure of the problem to efficiently discover high performing grasps for each object stable pose. \algabbr can be used to complement any general-purpose grasping algorithm with any grasp modality (parallel-jaw, suction, multi-fingered, etc) to learn policies for objects in which they exhibit persistent failures. Simulation experiments suggest that \algabbr can significantly outperform both general-purpose grasping pipelines and two other online learning algorithms and achieves performance within 5\% of the optimal policy within 1000 and 8000 timesteps on average across 46 challenging objects from the Dex-Net adversarial and EGAD! object datasets, respectively. Initial physical experiments suggest that \algabbr can improve grasp success rate by 45\% over a Dex-Net baseline with just 200 grasp attempts in the real world. See \url{https://tinyurl.com/exp-grasping} for supplementary material and videos.
\end{abstract}



\section{Introduction}
Robot grasping systems have a broad array of applications in manufacturing, warehousing, assistive robotics, and household automation~\cite{morrison2020learning, kalashnikov2018qt, levine2018learning, mahler2019learning}. There has been significant recent work in analytic grasp planning algorithms~\cite{bicchi2000robotic, murray2017mathematical, rimon2019mechanics, kim2013physically}, but these methods can often be difficult to apply when object geometry is unknown. In recent years, robot learning has shown exciting promise by utilizing large-scale datasets of previously attempted grasps both in simulation~\cite{mahler2019learning, kappler2015leveraging, morrison2020learning, viereck2017learning} and in physical experiments~\cite{pinto2016supersizing,levine2018learning,morrison2020learning,choi2018learning} to learn general-purpose grasping policies that can generalize to objects of varying geometries. To further enable generalization, there has also been work on applying reinforcement learning to learn grasping policies~\cite{kalashnikov2018qt, levine2018learning,breyer2019comparing,kroemer2019review}. However, while these techniques have shown significant success in practice, their generality comes at a cost: a policy which aims to grasp every object may fail to generalize to specific challenging objects~\cite{wang2019adversarial, memory-grasping}.

Inspired by infants that repeatedly attempt to grasp a toy until they can learn reliable ways to grasp it, we consider a novel problem: \probname, where a robot is presented with an unknown object and learns to reliably grasp it by repeatedly attempting grasps and allowing the object pose to evolve based on grasp outcomes. The objective is for the robot to explore grasps across different object poses to reliably grasp the object from any of its stable resting poses. We then present an algorithm for \probname motivated by polyhedral objects with sparse grasps and adversarial geometries, which can cause persistent failures in general-purpose grasping systems~\cite{memory-grasping, wang2019adversarial}. The intuition is that while a general-purpose grasping policy can be broadly applied to a large set of objects, it can be complemented by learned policies for specific challenging objects.

This paper contributes (1) \probname: an MDP formulation of the problem of discovering robust grasps across object poses through online interaction that is independent of the grasping modality (parallel-jaw, suction, multi-fingered, etc), (2) parameter-dependent performance bounds on a family of existing tabular reinforcement learning algorithms in this setting, (3) an efficient bandit-inspired algorithm, \algname, for \probname, and associated no-regret guarantees, (4) simulation experiments suggesting that \algabbr can significantly outperform baseline algorithms which explore via reinforcement learning or select actions using general-purpose grasping policies across 46 objects in both the Dex-Net adversarial and EGAD! object datasets, and (5) initial physical results suggesting that \algabbr can improve grasp success rate by 45\% on average over two challenging objects compared to a Dex-Net baseline.

\begin{figure*}[th!]
    \vspace{-2.5mm}
    \centering
    \includegraphics[width=\linewidth]{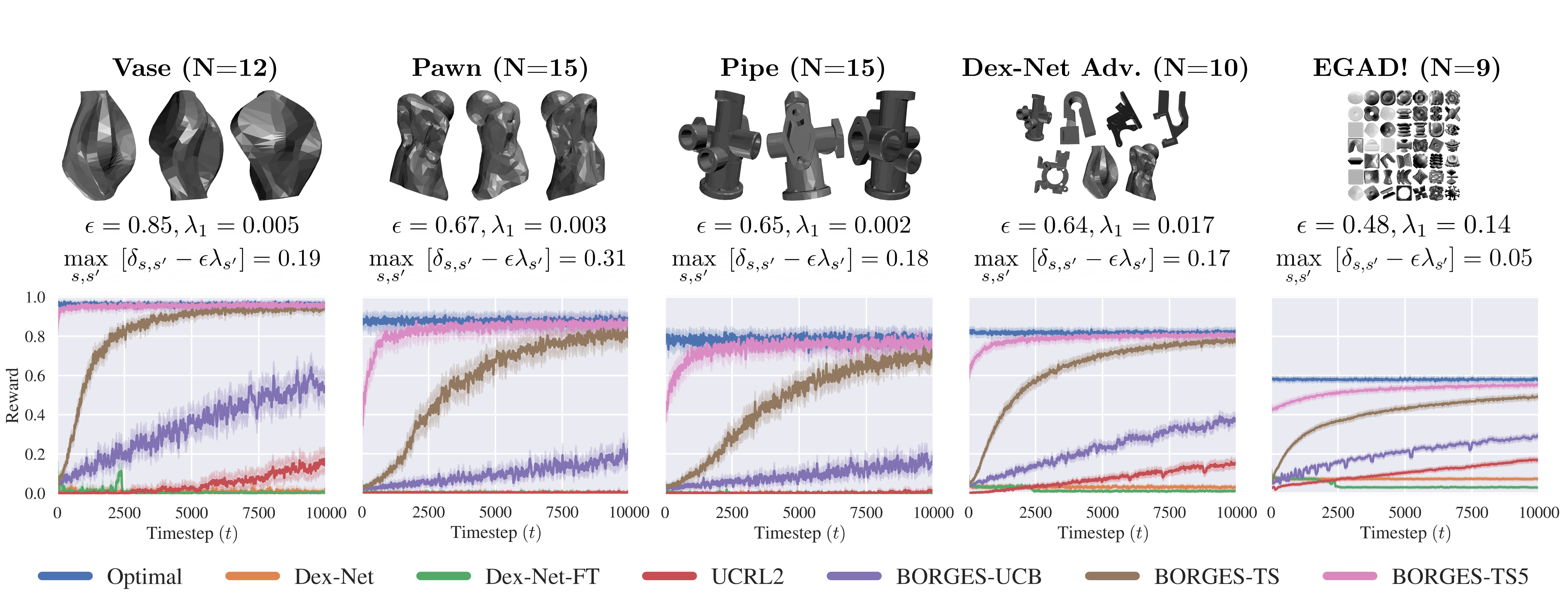}
    \caption{\textbf{Simulation Experiments: }Performance of each policy across the vase, pawn and pipe objects from the Dex-Net 2.0 adversarial object set~\cite{mahler2017dex} (first three columns), as well as aggregated performance over 7 Dex-Net objects (fourth column) and 39 EGAD! objects (fifth column). We report the number of distinguishable stable poses ($N$) with the first three shown in the second row, as well as $\epsilon$ (ground truth value for the lowest quality best grasp across poses), $\lambda_1$ (least likely stable pose probability) and $\max_{s,s'} \left[\delta_{s,s'} - \epsilon \lambda_{s'}\right]$ (maximum difference in transitioning via toppling and transitioning via grasping to a new pose) values. For the datasets, we give mean values and show the most likely stable pose of each object. We visualize policy performance in the learning curves in the lowest row. \algabbr quickly converges to near-optimal performance even when Assumption~\ref{assum:topple-bound} is violated while the other algorithms fail to reach optimal performance.
    \vspace{-2.5mm}}
    \label{fig:policy_rewards}
\end{figure*}

\section{Related Work}
Work in analytic robot grasping assumes knowledge of object geometry and pose to design geometric grasp planning algorithms~\citep{bicchi2000robotic, rimon2019mechanics, kim2013physically, murray2017mathematical}. Recently, learning-based general-purpose algorithms have emerged for planning grasps robust to uncertainty in object geometry and sensing on a wide range of objects with data-driven strategies~\citep{mahler2019learning,pinto2016supersizing,viereck2017learning, morrison2020learning, lenz2015deep, kappler2015leveraging} and online exploration through reinforcement learning~\citep{kalashnikov2018qt, levine2018learning}. The latter approaches have been very effective in learning end-to-end policies for grasp planning for a variety of objects. In contrast, we focus on learning policies for the specific challenging objects that pose problems for these systems~\citep{wang2019adversarial, memory-grasping}.

Significant prior work studies multi-armed bandit frameworks for online grasp exploration~\citep{laskey2015budgeted, laskey2015multi, oberlin2018autonomously, TSLP, kroemer2010combining, eppner2017visual, lu2020multi}, but primarily focus on settings when some geometric knowledge is known or grasp exploration is limited to a single object pose. In contrast, we consider a formulation where the robot must learn grasps across all poses of the object without human supervision. \citet{laskey2015multi} consider the setting where some prior geometric knowledge is known, but present an algorithm for 2D objects that does not use visual inputs.~\citet{TSLP} and \citet{oberlin2018autonomously} relax these assumptions by exploring grasps for a single stable pose of 3D object with RGB or depth observations. We extend these ideas by repeatedly dropping the object and exploring grasps in all encountered stable poses. Thus, \algabbr naturally explores grasps over the distribution of possible stable poses~\citep{goldberg1999part} to learn a robust policy which can reliably grasp the object when randomly dropped in the workspace.

A key requirement for successful exploration of grasps on an object is exploring the space of its resting stable poses, since the object will necessarily be in one of these poses during grasp attempts. There has been significant prior work on orienting parts into specific stable poses through a series of parallel jaw gripper movements~\citep{orienting-parts}, toppling actions~\citep{correa2019toppling}, and squeezing actions~\citep{goldberg1990bayesian}. However, these approaches require knowledge of object geometry apriori. When object geometry is not known, but assumed to be polyhedral, prior work~\citep{goldberg1999part} has established that repeatedly dropping the object from a known initial distribution of poses onto a flat workspace results in a stationary distribution over stable poses. This dropping procedure provides a useful primitive for reaching new stable poses of an object. We leverage this insight to formulate the \probname problem and design algorithms that address this setting to discover high-quality grasps for different object stable poses.
\section{\probname: Problem Statement}
\label{sec:framework}
Given a single unknown polyhedral object on a planar workspace, the objective is to learn a grasping policy that maximizes the likelihood of grasp success over all stable poses of the object~\citep{goldberg1999part,moll2002manipulation}. We formulate \probname as a Markov Decision Process (Section~\ref{subsec:formulation}), define assumptions on the environment (Section~\ref{subsec:assumptions}) and formulate the policy learning objective (Section~\ref{subsec:learning-obj}).

\subsection{\probname as an MDP}
\label{subsec:formulation}
We consider exploring grasps on an unknown, rigid, polyhedral object $\mathcal{O}$ which rests in one of a finite set of $N$ distinguishable stable poses with associated landing probabilities $\{\lambda_s\}_{s=1}^{N}$. We consider polyhedral objects since they admit a finite number of stable resting poses, as does any object defined with a triangular mesh~\citep{algebraic-parts}. We assume that the robot does not initially know any of these stable poses or their count $N$. The robot must discover new stable poses from experience by attempting grasps on the object in its current stable pose and lifting and then releasing the object when grasps are successful. We assume an overhead depth camera with known camera intrinsics that cannot reliably determine the 3D shape of the object, but can be used to recognize distinguishable 3D poses by performing planar translations and rotations of the image into a canonical orientation and translation. We also assume that the camera can be used after grasp attempts to determine if a grasp succeeds.

We model \probname as an MDP $\mathcal{M} = (\mathcal{S}, \mathcal{A}, P, R)$ as follows. We first define a one-to-many mapping from the set of object stable poses $\Sigma$ to the set of overhead point clouds $\mathcal{I}$ that are scale-invariant, translation-invariant, and rotationally-invariant about the vertical axis. Then, we define the state space as the set of \textit{distinguishable} stable poses $\mathcal{S}$, where $\mathcal{S}$ is the set of equivalence classes within $\Sigma$, where two poses are equivalent if they map to the same set of overhead point clouds $\mathcal{I}$. For example, all poses of a cube would map to the same overhead point cloud, so a cube only has 1 distinguishable stable pose. We assume point cloud $\mathcal{I}$ is obtained by deprojecting a depth image observation $o \in \mathbb{R}^+_{H \times W}$ taken from the overhead camera with known intrinsics. We define a set $\mathcal{A}_s$ of $K$ grasp actions, such as parallel-jaw or suction grasps, for each stable pose $s \in \mathcal{S}$ of the object. The $K$ grasps are sampled from $o$ for each stable pose as in~\citet{mahler2017dex}. The full action space is the union of the grasp actions available on each pose: $\mathcal{A} =
\underset{s \in \mathcal{S}}{\bigcup} \mathcal{A}_s$. We assume that the robot acts in an environment with unknown transition probability distribution $P(s' \ |\ s, a)$, which denotes the probability of the object transitioning to pose $s'$ if grasp action $a$ is executed in pose $s$. If $a$ is a successful grasp, then the object is released onto the workspace to sample a new stable pose $s'$ based on unknown landing probabilities $\{\lambda_s\}_{s=1}^{N}$, while if $a$ is a failed grasp, the object topples into some new pose $s'$ with unknown probability $\delta_{s, s'}$. The robot receives reward $R(s, a)$ for selecting grasp action $a$ on stable pose $s$, so $R(s, a) = 1$ if executing $a$ in stable pose $s$ results in the object being successfully grasped and lifted, and $0$ otherwise. We assume that rewards are drawn from a Bernoulli distribution with unknown parameter $\phi_{s,a}$: $R(s, a) \sim \textrm{Ber}(\phi_{s,a})$ for $a \in \mathcal{A}_s$.

\subsection{Assumptions}
\label{subsec:assumptions}
To study \probname, we first establish assumptions on the system dynamics to ensure that all poses are reachable and which describe how the object pose can evolve when the object is (1) released onto the workspace or (2) when a grasp is attempted. We make no assumptions on the grasping modality (e.g., parallel jaw, suction, multi-fingered, etc).
\begin{assumption}
\label{assum:grasp-dynamics}
\textbf{Grasp Dynamics:} If a grasp succeeds, we assume that the robot can randomize the pose of the object before releasing it to sample subsequent stable poses from the associated unknown stable pose landing probabilities $\{\lambda_s\}_{s=1}^{N}$ for $\mathcal{O}$. If a grasp fails, we assume that the object's pose will either remain unchanged or topple into some other pose $s'$ with unknown probability $\delta_{s, s'}$.
\end{assumption}
\begin{assumption}
\label{assum:drop-dynamics}
\textbf{Release Dynamics:} 
The categorical distribution over stable poses defined by landing probabilities $\{\lambda_s\}_{s=1}^{N}$ is stationary and independent of prior actions and poses when $\mathcal{O}$ is released from a fixed height with its orientation randomized as in~\citet{goldberg1999part}. 
\end{assumption}
\begin{assumption}
\label{assum:ergodicity}
\textbf{Irreducibility: } We assume that there exists a policy $\pi$ such that the Markov chain over stable poses induced by executing $\pi$ in $\mathcal{M}$ is irreducible, and thus can reach all stable poses with nonzero probability for any initialization. 
\end{assumption}

Note that Assumption~\ref{assum:ergodicity} is satisfied if, for all poses $s \in \mathcal{S}$, $\lambda_s > 0$ and there exists a grasp with success probability $\epsilon > 0$. We assume that these conditions hold for analysis, but since object toppling is also possible, note that these conditions are sufficient but not necessary for ensuring irreducibility.

\subsection{Learning Objective}
\label{subsec:learning-obj}
The objective is to learn a policy $\pi: \mathcal{S} \rightarrow \mathcal{A}$ that maximizes the expected average reward over an infinite time horizon under the state distribution induced by $\pi$. Let $\tau = \{ (s_t, \pi(s_t)) \}_{t=1}^{T}$ be a trajectory of all states and actions when executing $\pi$ for time $T$ and let $r(\tau) = \sum_{t=1}^{T} R(s_t, \pi(s_t))$ be the sum of rewards for all states and actions in $\tau$, and let $p(\tau | \pi)$ be the trajectory distribution induced by policy $\pi$. Then the expected average reward obtained from policy $\pi$ in $\mathcal{M}$ is given as:
\begin{align}
    \label{eq:policy-return}
    J(\mathcal{M}, \pi, T) = \frac{1}{T} \mathbb{E}_{\tau \sim p(\tau |\pi)} \left[r(\tau)\right]
\end{align}
The objective is to find the policy which maximizes expected reward over an infinite time horizon:
\begin{align}
    \label{eq:policy-learning-obj}
    \pi^* = \argmax_{\pi} \lim_{T \rightarrow \infty} J(\mathcal{M}, \pi, T)
\end{align}

\section{Reinforcement Learning for \probname}
\label{sec:rl}
We first consider the performance of reinforcement learning algorithms for \probname and leverage the structure of the MDP described in Section~\ref{sec:framework} to establish a bound on the cumulative regret for a family of existing tabular reinforcement learning algorithms when applied to \probname. See Section~\ref{sec:proofs} of the supplementary material for all proofs.

\subsection{Analyzing the \probname MDP}
\label{subsec:analyze-MDP}
A common metric to measure policy performance in online-learning settings is \textit{regret}, which has been analyzed in the reinforcement learning setting by a variety of prior work~\cite{UCRL2, UCRLV, KL-UCRL, PSRL}. Intuitively, regret quantifies the difference in accumulated reward within $T$ timesteps between a given policy $\pi$ and optimal infinite horizon policy $\pi^*$ for MDP $\mathcal{M}$. More precisely, we define average regret based on the definition in~\cite{UCRL2}:
\begin{align}
    \textrm{Regret}(\mathcal{M}, \pi, T) = \max_{\pi'} \left[\lim_{T \rightarrow \infty} J(\mathcal{M}, \pi', T)\right] - J(\mathcal{M}, \pi, T)
\end{align}

Recent theoretical work by~\citet{UCRL2} on reinforcement learning for tabular MDPs yielded algorithms which can attain average regret proportional to the \emph{diameter} of the MDP, a measure of the furthest distance between pairs of states under an appropriate policy. However, for general MDPs, this diameter can be arbitrarily large, making these regret bounds difficult to interpret in practical settings. We leverage the structure of the MDP to derive an upper bound on the MDP diameter, which precisely quantifies the difficulty of grasp exploration based on the parameters of $\mathcal{M}$.

We begin by defining the Markov chain over $\mathcal{S}$ induced by a stationary deterministic policy $\pi$.
\begin{definition}
\label{def:markov-chain}
\textbf{Pose Evolution under $\pi$: } Given stationary policy $\pi$, the transitions between pairs of states in $\mathcal{M}$ is defined by a Markov chain. Precisely, the transition probabilities under $\pi$, denoted by $P^{\pi}$ where $P^{\pi}[s,s'] = P(s' \ |\ s, a = \pi(s))$, are given as follows:
\begin{align}
P^{\pi}[s, s'] =  \phi_{s,\pi(s)} \lambda_{s'} + (1 - \phi_{s,\pi(s)}) \delta_{s, s'}
\end{align}
\end{definition}
Given this Markov Chain over poses for $\pi$, we can now analyze the diameter of the MDP, denoted $D(\mathcal{M})$, by considering the hitting time between stable poses in $\mathcal{M}$ as defined in~\cite{hit-time}.
\begin{definition}
    \label{def:diam}
    Let $T^\pi_{s \rightarrow s'}$ denote the expected hitting time between states $s$ and $s'$ under policy $\pi$ under the Markov chain defined in Definition~\ref{def:markov-chain}. Then the diameter of $\mathcal{M}$ is defined as follows~\cite{UCRL2}:
    \begin{align}
        D(\mathcal{M}) = \max_{s \neq s'} \min_{\pi} \mathbb{E}\left[T^\pi_{s \rightarrow s'}\right]
    \end{align}
\end{definition}
Intuitively, $D(\mathcal{M})$ measures the temporal distance between the most distant states in an MDP under the policy that minimizes this distance. We now leverage the structure of $\mathcal{M}$ to establish an upper bound on $D(\mathcal{M})$.
\begin{lemma}
    \label{lemma:diam-bound}
    The diameter of $\mathcal{M}$ can be bounded above as follows:
    \begin{align}
        D(\mathcal{M}) \leq \frac{1}{\epsilon \lambda_1}
    \end{align}
    Here $\epsilon$ is a lower bound on the success probability of the highest quality grasp over all stable poses and $\lambda_1$ is the landing probability for the least likely stable pose.
\end{lemma}
Lemma~\ref{lemma:diam-bound} captures the intuition that the diameter of the MDP is large if the best grasp in each stable pose has a low success probability ($\epsilon$ is small), or if there exists a stable pose with very low landing probability ($\lambda_1$ is small).

Now  we can establish regret bounds for $4$ tabular reinforcement learning algorithms (UCRL2~\cite{UCRL2}, KL-UCRL~\cite{KL-UCRL}, PSRL~\cite{FMDP}, and UCRLV~\cite{UCRLV}) when applied to $\mathcal{M}$ by combining diameter dependent regret bounds from prior work and the bound on the diameter of $\mathcal{M}$ established in Lemma~\ref{lemma:diam-bound}.
\begin{theorem}
    \label{thm:regret-bound}
    UCRL2~\cite{UCRL2}, KL-UCRL~\cite{KL-UCRL} and PSRL~\cite{FMDP} achieve average regret given by $\text{Regret}(\mathcal{M}, \pi, T) \sim \tilde{O} \left( \frac{N}{\epsilon\lambda_1} \sqrt{\frac{K}{T}}\right)$ for any \probname MDP $\mathcal{M}$ while UCRLV~\cite{UCRLV} achieves average regret given by $\text{Regret}(\mathcal{M}, \pi, T) \sim \tilde{O} \left( \sqrt{\frac{NK}{\epsilon\lambda_1 T}}\right)$. Here $N$, $K$, $T$, $\epsilon$ and $\lambda_1$ are defined as in Section~\ref{sec:framework}.
\end{theorem}

Theorem~\ref{thm:regret-bound} leverages the specific structure of the MDP in \probname to relate the accumulated regret of $4$ tabular RL algorithms to intuitive parameters of the MDP, providing insight into the theoretical complexity of \probname in the context of reinforcement learning.

\section{\algname}
\label{sec:bandits}
While the reinforcement learning algorithms in Section~\ref{sec:rl} provide a method and formal guarantees for \probname, the algorithms themselves do not exploit any specific structure in MDP $\mathcal{M}$. One key feature of \probname is that in practice, objects have a small, finite set of stable poses~\cite{goldberg1999part}. This motivates learning a set of $N$ policies, each of which explore grasps in a particular object stable pose. However, the grasp exploration problem in each pose is coupled. For example, there may exist a pose $s$ with no available high quality grasps but a high likelihood of a failed grasp causing the object to topple into another pose with high quality grasps. Then, the optimal policy may deliberately fail to grasp the object in pose $s$ in order to obtain access to grasps in the more favorable stable pose, leading it to avoid grasp exploration in poses without high quality grasps. To avoid this behavior, we introduce Assumption~\ref{assum:topple-bound}.
\begin{assumption}
\label{assum:topple-bound}
    We assume that $\delta_{s,s'} \leq \epsilon \lambda_{s'}$ for all $s \neq s'$ where $\delta_{s,s'}$ is the probability of toppling into pose $s'$ given a failed grasp in pose $s$, $\epsilon$ is a lower bound on the success probability of the highest quality grasp over all stable poses as defined above, and $\lambda_{s'}$ is the landing probability of pose $s'$.
\end{assumption}
Assumption~\ref{assum:topple-bound} ensures that there exists a grasp in all stable poses $s$ such that the probability of transitioning to new pose $s'$ via a grasp attempt is higher than that of toppling from pose $s$ to pose $s'$. Given this assumption, the optimal grasp exploration policy in $\mathcal{M}$ reduces to selecting the grasp with highest success probability in each encountered pose, as this policy maximizes both reward at the current timestep and exploration of other stable poses when the object is released. In other words, the global optimal policy is the \textit{greedy} policy. Given this structure, we can view the grasp exploration problem as $N$ independent multi-armed bandit problems corresponding to grasp exploration in each pose. However, although grasp exploration can be performed independently in each pose, the success of a grasp exploration policy in one pose affects the time available to explore grasps in another pose, thereby coupling each bandit problem.

We propose an algorithm that takes advantage of this structure to enable rapid online grasp exploration: \algname\footnote{Jorge Luis Borges (1899-1986) was a brilliant writer whose short stories considered geometry, time, and combinatorics.}. \algabbr works by maintaining the parameters of $N$ independent bandit policies $(\pi^{\mathcal{B}}_s)_{s=1}^{N}$, where $\pi^{\mathcal{B}}_s: s \rightarrow \mathcal{A}_s$ and $\pi^{\mathcal{B}}_s$ is only \textit{active} in pose $s$. Let $\pi^{\mathcal{B}}$ denote the meta-policy induced by executing $\pi^{\mathcal{B}}_s$ in pose $s$ and assume that $\pi^{\mathcal{B}}_s$ is learned by running a no-regret online learning algorithm $\mathcal{B}$ for grasp exploration in $s$. No-regret algorithms for the stochastic multi-armed bandit problem include the UCB-1 algorithm~\cite{auer2002finite} and Thompson Sampling~\cite{TS-regret}. We then formulate a new notion of regret capturing the gap between $\pi^{\mathcal{B}}$ and the optimal policy on their respective distributions and show that \algabbr achieves vanishing average regret despite the interdependence between pose exploration times. 

Let $p^{\mathcal{B}}_T$ denote the distribution of poses seen under the sequence of policies $\pi^{\mathcal{B}}_{1:T}$ at each round of learning up to time $T$ and let $p^*_T$ denote the distribution of poses seen when executing the optimal policy ($\pi^*$) in $\mathcal{M}$ up to time $T$. We define the average regret achieved by \algabbr in MDP $\mathcal{M}$ after $T$ rounds as the difference in accumulated reward of the optimal policy on pose distribution $p^*_T$ and the accumulated reward of $\pi^{\mathcal{B}}$ on pose distribution $p^{\mathcal{B}}_T$.
\begin{definition}
    \label{def:avg-bandit-regret}
    The average regret accumulated by running $\mathcal{B}$ in each stable pose is defined as the difference between the average regret on each pose visited by the optimal policy $\pi^*$ weighted by the probability of it visiting each pose and the corresponding quantity for the executed policy $\pi^\mathcal{B}$:
    \begin{align*}
          \mathbb{E}\left[\mathcal{R}^{\mathcal{B}}(T)\right] &= \sum_{s=1}^{N} p^*_T(s) \mathbb{E}\left[\frac{1}{T_s^*}\sum_{t=1}^{T^*_s} R(s, \pi^* (s)) \right] - 
          \sum_{s=1}^{N} p^{\mathcal{B}}_T (s) \mathbb{E}\left[\frac{1}{T_s^\mathcal{B}}\sum_{t=1}^{T^\mathcal{B}_s} R(s, \pi^{\mathcal{B}}_{t} (s)) \right]
    \end{align*}
    where $T_s^*$ is the time spent by the optimal policy in pose $s$ and $T_s^{\mathcal{B}}$ is the time spent by $\pi^{\mathcal{B}}$ in pose $s$.
\end{definition}
In Section~\ref{sec:proofs} of the supplementary material, we show that the average regret as defined in Definition~\ref{def:avg-bandit-regret} vanishes to $0$ in the limit as $T \rightarrow \infty$ as stated in Theorem~\ref{thm:avg-bandit-regret}.
\begin{theorem}
    \label{thm:avg-bandit-regret}
    The average regret achieved by \algabbr, when using any no-regret bandit algorithm $\mathcal{B}$ for grasp exploration in each encountered stable pose, vanishes in the limit:
    \begin{align*}
        \underset{T \rightarrow \infty}{\lim} \mathbb{E}\left[\mathcal{R}^{\mathcal{B}}(T)\right] = 0
    \end{align*}
\end{theorem}
This result leverages the precise structure of $\mathcal{M}$, namely that the optimal policy is the greedy policy, to provide sublinear regret guarantees for \algabbr.

A convenient property of \algabbr is that while exploring grasps on an object, it naturally explores different object poses as well. Theorem~\ref{theorem:pose-coverage} establishes an upper bound on the expected number of timesteps required for \algabbr to reach all object stable poses.
\begin{theorem}
    \label{theorem:pose-coverage}
    Let $T_{\text{cover}}$ denote the number of grasps \algabbr executes until it has reached every stable pose. We can bound $\mathbb{E}\left[T_{\text{cover}}\right]$ above as follows where $K$, $\epsilon$, and $\lambda$ are defined as in Section~\ref{sec:framework}.
    \begin{align*}
        \mathbb{E}\left[T_{\text{cover}}\right] \leq \frac{K}{\epsilon}\sum_{j=1}^N \frac{1}{1 - \sum_{i=2}^{j} \lambda_{i-1}}
    \end{align*}
\end{theorem}

We note that there can be pathological objects for which Assumption~\ref{assum:ergodicity} is violated due to a sink state from which the object can neither be grasped nor toppled into a new pose. In this case, no grasp exploration strategy will be successful, because there will exist poses from which the robot cannot recover. However, one benefit of \algabbr is that it can detect these cases, as it maintains estimates of the success probabilities of each grasp during learning and thus could request external intervention from a human supervisor when it becomes clear that a given pose is a sink state in $\mathcal{M}$. We also acknowledge that many objects violate Assumption~\ref{assum:topple-bound} as they can be easily toppled between poses, including several of the objects we use for evaluation. However, we find that even when Assumption~\ref{assum:topple-bound} is violated, \algabbr still achieves good performance in practice (Figure~\ref{fig:policy_rewards}).
\section{Experiments}
\subsection{Simulation Experiments}
\label{sec:sim-exps}
In simulation experiments, we study whether \algabbr can discover high quality grasps on objects for which general-purpose grasping policies, such as Dex-Net 4.0~\cite{mahler2019learning} and GG-CNN~\citep{morrison2020learning}, perform poorly and whether learning different policies for each stable pose accelerates grasp exploration.

\textbf{Simulation Details: }To evaluate each policy, we choose 7 objects from the set of Dex-Net 2.0 adversarial objects~\cite{mahler2017dex} and 39 evaluation objects from the EGAD! dataset~\citep{morrison2020egad} for which general-purpose parallel-jaw grasping policies (Dex-Net 4.0~\cite{mahler2019learning} and GG-CNN~\cite{morrison2020learning}) perform poorly, despite there existing high-quality grasps in multiple stable poses. We sample a set of $K$ parallel-jaw grasps on the image observation of pose $s$ of each object as in~\cite{mahler2017dex}, and calculate the ground-truth quality of each grasp, $\phi_{s, a}$, using a wrench resistance metric that measures the ability of the grasp to resist gravity~\cite{mahler2018dex}. We randomize the initial pose of each object and execute \algabbr and baselines, sampling rewards from $\text{Ber}(\phi_{s, a})$. If the grasp succeeds, we randomize the pose, choosing a stable pose according to the stable pose distribution of the object. Otherwise, the object may topple into a new stable pose. We execute each policy for 10 rollouts of 10 trials, where new grasps are sampled for each trial and each rollout evaluates the policy over 10,000 timesteps of grasp exploration. Since there is stochasticity in both the grasp sampling and the policies themselves, we average policy performance across the 10 rollouts and 10 trials. In addition, we smooth policy performance across a sliding window of 20 timesteps and report average reward for each timestep. 

To model object toppling when grasps are unsuccessful (Section~\ref{subsec:assumptions}), we use the toppling analysis from \citet{correa2019toppling} to determine the toppling transition matrix for a given object. Specifically, we generate the distribution of next states from a given state by sampling non-colliding pushes across vertices on the object, finding their distribution of next states given perturbations around the nominal push point, and average the distribution from all of the pushes. Then if a grasp fails, we choose the next state according to the corresponding topple transition probabilities. When evaluating policies, we remove poses that have no grasps with nonzero ground-truth quality and renormalize the stable pose distribution. We emphasize that we perform this step for all policies and do this only to ensure that no poses act as sink states from which the robot can never change the pose in the future (and thus ensure that Assumption~\ref{assum:ergodicity} is satisfied). For the 7 objects from the Dex-Net 2.0 adversarial object dataset, $11\%$ of poses (an average of 1.3 poses per object) were removed. We present additional experiments in the supplementary material where this pose removal step is not performed, and find that while all policies perform more poorly (Section~\ref{sec:additional-exps}), \algabbr still outperforms all baseline policies. We also include further details on the experimental setup in the supplementary material.

\textbf{Policies: }We compare 3 variants of \algabbr against 3 baselines to evaluate whether \algabbr is able to (1) substantially outperform general-purpose grasping policies~\citep{mahler2019learning} on challenging objects and (2) learn more efficiently than other online learning algorithms which update a grasp quality convolutional network (GQCNN) online or explore grasps via reinforcement learning. We also instantiate \algabbr with different algorithms for $\mathcal{B}$ to study how this choice affects grasp exploration efficiency and implement an oracle baseline to study whether \algabbr is able to converge to the optimal policy. We compare the following baselines and \algabbr variants: \textbf{Dex-Net}, which selects grasps greedily  with respect to the Grasp Quality Convolutional Network from Dex-Net 4.0~\cite{mahler2019learning} with probability $0.95$ and selects a grasp uniformly at random with probability $0.05$, \textbf{Dex-Net-FT}, which additionally fine tunes the GQCNN policy online from agent experience, an implementation of \textbf{UCRL2} from~\cite{UCRL_github}, a tabular RL algorithm discussed in Section~\ref{sec:rl}, and instantiations of \algabbr with the UCB-1 algorithm~\cite{auer2002finite} (\textbf{\algabbr-UCB}), Thompson sampling (\textbf{\algabbr-TS}) with a uniform Beta distribution prior, and Thompson sampling with a prior from the Dex-Net policy of strength 5 (\textbf{\algabbr-TS5}) as in~\cite{TSLP}. Finally, we implement an oracle baseline that chooses grasps with the best ground-truth metric at each timestep to establish an upper bound on performance.

\textbf{Simulation Results: }Results for baselines and \algabbr variants are in Figure~\ref{fig:policy_rewards}. Above the learning curves in Figure~\ref{fig:policy_rewards}, we report the maximum violation of Assumption~\ref{assum:topple-bound} across all pose pairs for each object ($\max_{s,s'} \left[ \delta_{s,s'} - \epsilon\lambda_{s'}\right]$). This value captures how much more likely an object is to topple between poses $s$ and $s'$ than be successfully grasped in $s$ and released into $s'$ for pose pair $(s, s')$ which most violates Assumption~\ref{assum:topple-bound}. \algabbr is robust to large violations of Assumption~\ref{assum:topple-bound}, and substantially outperforms prior methods even when there exist pose pairs where it is $30\%$ more likely to topple between the poses than transition between them with a successful grasp.

As shown in Figure~\ref{fig:policy_rewards}, the Dex-Net policy typically performs very poorly, achieving an average reward of less than 0.1 per timestep. While the online learning policies also start poorly, they quickly improve, and \algabbr-\textsc{TS} eventually converges to the optimal policy. We also find that \algabbr-\textsc{TS5}, which leverages Dex-Net as a prior using the method presented in~\cite{TSLP}, further accelerates convergence to the optimal policy. This result is promising, as it suggests that the exploration strategy in \algabbr can be flexibly combined with general-purpose grasping policies to significantly accelerate grasp exploration on unknown objects. We find that the Dex-Net-FT policy performs very poorly even though it continues to update the weights of the network online with the results of each grasp attempt and samples a random grasp with probability $0.05$, which aids exploration. We hypothesize that this is due to the initial poor performance of Dex-Net---since the vast majority of fine-tuning grasps attain zero reward, the network is unable to explore enough high-quality grasps on the object. Overall, these results suggest that \algabbr can greatly increase grasp success rates on objects for which Dex-Net performs poorly. We also find that \algabbr policies greatly outperform tabular RL policy UCRL2 by leveraging the structure of the MDP. Both the UCB and Thompson sampling implementations maintain separate policies in each pose, leveraging the fact that the optimal policy is the greedy policy. This allows the \algabbr policies to not waste timesteps exploring possible transitions to poses with low rewards. Thompson sampling additionally leverages the fact that the rewards are distributed as Bernoulli random variables, which may explain the significant performance gap between the Thompson sampling and UCB implementations. Thus, \algabbr policies quickly learn to choose high-quality grasps in each pose to transition quickly to new poses.

We perform sensitivity analysis of \algabbr to $\epsilon$ and $\lambda_1$ and find that \algabbr quickly converges to the optimal policy unless $\epsilon$ or $\lambda_1$ is low. We additionally evaluate \algabbr for different numbers of sampled grasps ($K$) on each pose, and find that with decreasing values of $K$, all policies perform worse since the likelihood of sampling a high quality grasp decreases, but \algabbr policies continue to outperform prior methods. See Section~\ref{sec:sensitivity-exps} of the supplementary material for more details.

\subsection{Initial Physical Experiments}
\label{sec:phys-exps}
\begin{wrapfigure}{r}{0.5\linewidth}
    \centering
    \includegraphics[width=\linewidth]{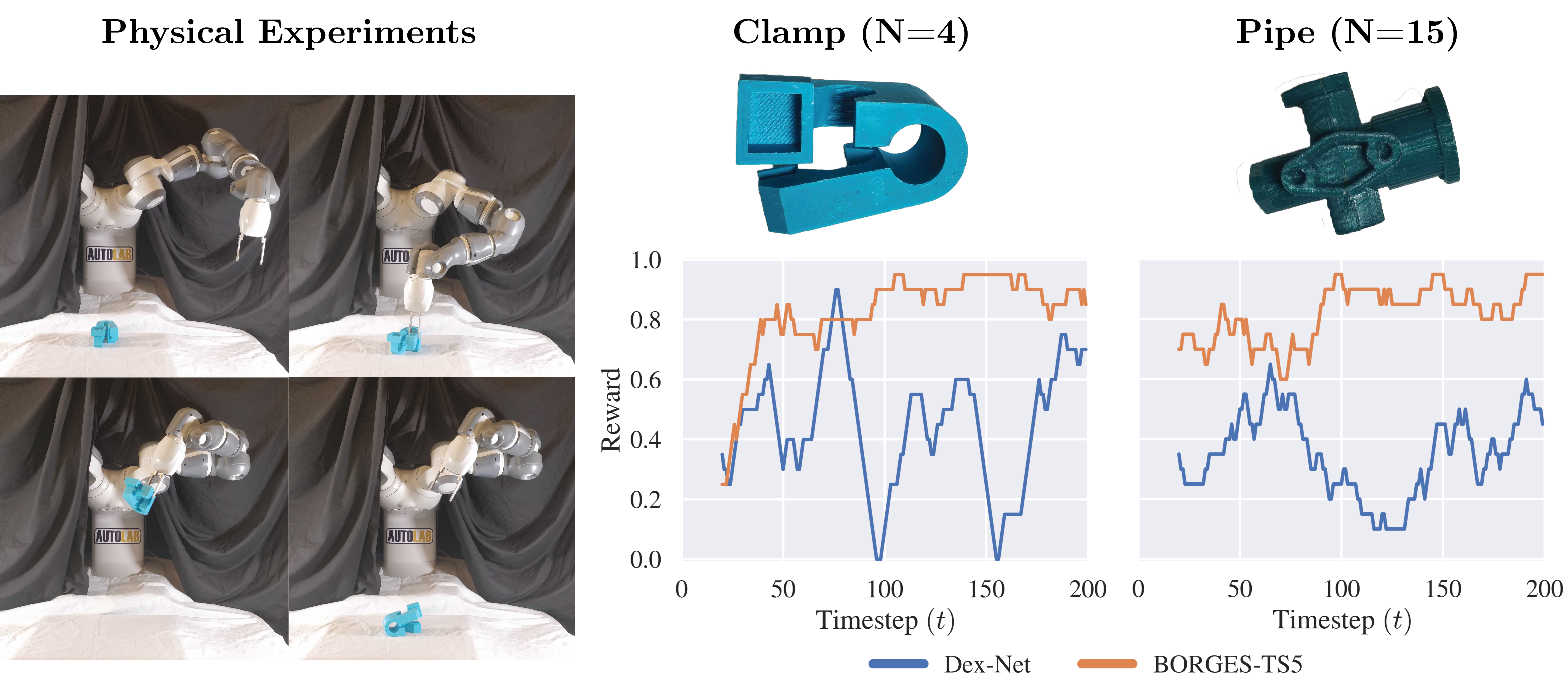}
    \caption{Experiment setup and learning curves for the Dex-Net and \algabbr-TS5 policies for two objects across 200 grasp attempts (smoothed with a running average of 20 attempts). The robot attempts to grasp the object at each timestep and, if it succeeds, rotates and drops the object to sample from the stable pose distribution (left). \algabbr-TS5 quickly converges within 100 attempts on both objects, indicating that it finds grasps that succeed nearly every time for each pose. Dex-Net's performance remains uneven, indicating that it finds high-quality grasps for some poses, but not others.}
    \label{fig:physical_rollouts}
\end{wrapfigure}

In physical experiments, we evaluate \algabbr on two challenging objects on an ABB YuMi robot and compare performance with the Dex-Net policy from Section~\ref{sec:sim-exps}.  Section~\ref{sec:phys-exps-supp} of the supplementary material contains further details on the experimental setup and procedures. Figure~\ref{fig:physical_rollouts} shows learning curves for both \algabbr and Dex-Net; results suggest that \algabbr quickly finds high-quality grasps across all object poses, converging to near-perfect performance. Dex-Net correctly predicts high-quality grasps on some of the poses, but attempts suboptimal grasps in others, resulting in high variance performance across the rollout since it is unable to adapt to its successes and failures. Across the final 100 timesteps for each object (i.e., after \algabbr begins to explore all poses), \algabbr reaches $0.89$ and $0.87$ success rates on the clamp and pipe, respectively, as compared to $0.49$ and $0.37$ for the Dex-Net policy after just 200 grasp attempts in real world. This highlights the importance of a policy that can learn online from successful and failed grasp attempts; Dex-Net does not learn online so continuing to attempt grasps will lead to the same high-variance behavior over time, but \algabbr continues to stabilize as it approaches optimal performance across all stable poses.

\section{Future Work}
In future work, we plan to evaluate \algabbr in extensive physical trials with different grasping modalities such as suction grasps and with priors from different grasp planners. One property of \algabbr is that in the process of exploring grasps, it also explores different object stable poses. Thus, we are excited to explore further applications of \algabbr; for example, exploring different object poses to construct accurate 3D models of unknown objects or inspect parts for defects.
\clearpage
\acknowledgments{This research was performed at the AUTOLAB at UC Berkeley in affiliation with the Berkeley AI Research (BAIR) Lab, and the CITRIS ''People and Robots" (CPAR) Initiative. This research was supported in part by the Scalable Collaborative Human-Robot Learning (SCHooL) Project, NSF National Robotics Initiative Award 1734633, by donations from Google, Toyota Research Institute, Siemens, Autodesk, Honda, Intel, Hewlett-Packard and by equipment grants from PhotoNeo, NVidia, and Intuitive Surgical.
Ashwin Balakrishna and Michael Danielczuk are supported by the National Science Foundation Graduate Research Fellowship Program under Grant No. 1752814. We thank our colleagues who provided helpful feedback, especially Brijen Thananjeyan and Jeffrey Ichnowski.}

\begin{small}
\bibliography{corl}  
\end{small}

\newpage
\appendix
\setcounter{page}{1}
\resetlinenumber
\setcounter{figure}{0}
\begin{LARGE}
\begin{center}
\textbf{\probname:\\Asymptotically Optimal Algorithms for \\ Grasping Challenging Polyhedral Objects \\Supplementary Material}
\end{center}
\end{LARGE}
\maketitle

The supplementary material is organized as follows. In Section~\ref{sec:proofs} we provide proofs for all theoretical results in Sections~\ref{sec:rl} and~\ref{sec:bandits}. In Section~\ref{sec:exp-details} we provide additional experimental details, in Section~\ref{sec:additional-exps} we include additional simulation experiments, and in Section~\ref{sec:sensitivity-exps} we perform sensitivity experiments. In Section~\ref{sec:phys-exps} we provide further details on physical experiments.

\section{Proofs}
\label{sec:proofs}
Here we provide the proofs for all results in Section~\ref{sec:rl} and Section~\ref{sec:bandits} in the main text.
\subsection{Proof of Lemma~\ref{lemma:diam-bound}}
\label{sec:lemma-proof}
Consider \probname MDP $\mathcal{M}'$ for which the object stable pose does not change when a grasp fails. Thus, it must be the case that $\delta_{l, m} = 0$ when $l \neq m$ and $\delta_{l, l} = 1 \ \forall l$. Note that for any \probname MDP $\mathcal{M}$, it must be the case that $D(\mathcal{M}) \leq D(\mathcal{M}')$ since the probability of transition between any pair of distinct states in $\mathcal{M}'$ is at most the probability of transition in $\mathcal{M}$. Now we establish a bound on $D(\mathcal{M})$ by bounding $D(\mathcal{M}')$.
 
Without loss of generality, let $\lambda_1 \leq \lambda_s \ \forall s \in \mathcal{S}$. Additionally, define $\pi^*$ as the policy which selects the grasp with highest success probability on all poses, with associated probability transition matrix $P^{\pi^*}$ and with hitting time $T^{\pi^*}_{s \rightarrow s'}$ defined as in Definition~\ref{def:markov-chain}.  Then, the diameter of $\mathcal{M}'$ can be computed as follows.

For MDP $\mathcal{M}'$, it must be the case that
\begin{align}
    \min_{\pi} T^\pi_{s \rightarrow s'} = T^{\pi^*}_{s \rightarrow s'} 
\end{align}
since $\pi^*$, the policy which always picks the highest quality grasp on each pose, minimizes the hitting time between any pair of poses $s, s'$. Furthermore, note that
\begin{align}
    \max_{s \neq s'}  T^{\pi^*}_{s \rightarrow s'} = \max_s T^{\pi^*}_{s \rightarrow 1} 
\end{align}
since for any starting pose $s$, the hitting time between $s$ and $s'$ will always be highest for $s' = 1$ (the pose with lowest drop probability) for any policy $\pi$. Thus, we see that 
\begin{align}
    \label{eq:diam-exact}
    D(\mathcal{M}') = \max_s T^{\pi^*}_{s \rightarrow 1}
\end{align}

Finally, we leverage equation~\ref{eq:diam-exact} to compute an upper bound on $D(\mathcal{M}')$ as follows.

Let $\pi_{\epsilon}$ be any policy which selects a grasp with success probability $\epsilon$ on each stable pose $l$. Note that by Assumption~\ref{assum:ergodicity} we assume that there exists a grasp with success probability at least $\epsilon$ on each pose. Without loss of generality, we can consider the case that there exists a grasp with success probability exactly $\epsilon$ on each pose and the policy which selects these grasps since the hitting times under this policy will only be greater than those under a policy which selects grasps with success probability greater than $\epsilon$. Then, it follows that \begin{align}
    \label{eq:hit-time-bound}
    \min_{\pi} T^\pi_{s \rightarrow s'} \leq T^{\pi_{\epsilon}}_{s \rightarrow s'} \ \forall s, s'
\end{align}
Now note that since we are considering pose evolution under $\pi_{\epsilon}$, which selects a grasp of the same quality $\epsilon$ on any pose, the starting pose $s$ does not affect the hitting time. Combining this with the fact that the hitting time to the least likely pose (pose $1$) will always be the highest for any policy $\pi$ and inequality~\eqref{eq:hit-time-bound} yields that 
\begin{align}
    D(\mathcal{M}') \leq \max_{s \neq s'} T^{\pi_{\epsilon}}_{s \rightarrow s'} = T^{\pi_{\epsilon}}_{2 \rightarrow 1}
\end{align}
Since the choice of $s$ does not matter under $\pi_\epsilon$, we use $s=2$ above without loss of generality. Now, note that the hitting time to pose $1$ under $\pi_\epsilon$ is distributed as a geometric random variable with parameter $\epsilon \lambda_1$, which has mean $\frac{1}{\epsilon \lambda_1}$, yielding the desired result. \qed

\subsection{Proof of Theorem~\ref{thm:regret-bound}}
\label{sec:diam-thm-proof}
The result immediately follows from combining the diameter bound from Lemma~\ref{lemma:diam-bound} and the regret bounds established for UCRL2~\cite{UCRL2}, KL-UCRL~\cite{KL-UCRL} and PSRL~\cite{FMDP} (average regret $\tilde{O}(DS \sqrt{A/T})$) and for UCRLV~\cite{UCRLV} (average regret $\tilde{O}( \sqrt{DSA/T})$) where $D$ is the MDP diameter and $S$ and $A$ are the cardinalities of the state space and action space respectively. \qed

\subsection{Proof of Theorem~\ref{thm:avg-bandit-regret}}
We bound the expected regret of \algabbr by decomposing the regret into two terms, one which depends on the divergence between the distribution of poses seen by the optimal policy and $\pi^{\mathcal{B}}$, and the other of which depends on the difference in rewards attained by $\pi^{\mathcal{B}}$ and $\pi^*$ when evaluated on the distribution of poses seen by $\pi^{\mathcal{B}}$. For simplicity, we refer to $\pi^{\mathcal{B}}$ as $\pi$ for the proof.
\label{sec:bandit-thm}
\begin{align}
    \mathbb{E}\left[\mathcal{R}^{\mathcal{B}}(T)\right] &= \sum_{s=1}^{S} p^*_T(s) \mathbb{E}\left[\frac{1}{T_s^*}\sum_{t=1}^{T^*_s} R(s, \pi^* (s)) \right] - 
    \sum_{s=1}^{S} p^{\pi}_T (s) \mathbb{E}\left[\frac{1}{T_s^\pi}\sum_{t=1}^{ T^\pi_s} R(s, \pi_{t} (s)) \right]\label{bandit-proof-1} \\
    &= \sum_{s=1}^{S} \left(p^*_T(s) - p^\pi_T(s) \right) g_s^*(T_s^*) +
    \sum_{s=1}^{S} p^{\pi}_T (s) \left( g_s^*(T_s^*) - g_s^\pi(T_s^\pi)\right)\label{bandit-proof-2} \\
    &= \mathbb{E} \left[ \mathcal{R}^\mathcal{B}_\pi (T)\right] + \sum_{s=1}^{S} \left(p^*_T(s) - p^\pi_T(s)\right) g_s^*(T_s^\pi)\label{bandit-proof-3}
 \end{align}
where~\eqref{bandit-proof-2} follows from letting $g_s^* (T_s^*) = \mathbb{E} \left[\frac{1}{T_s^*}\sum_{t=1}^{T_s^*} R(s, \pi^*(s))\right]$ and $g_s^\pi (T_s^\pi) = \mathbb{E} \left[ \frac{1}{T_s^\pi}\sum_{t=1}^{T_s^\pi} R(s, \pi_t(s)) \right]$ and~\eqref{bandit-proof-3} follows from letting $\mathbb{E} \left[ \mathcal{R}^\mathcal{B}_\pi (T)\right]$ denote the expected regret on the distribution of poses visited by policy $\pi$ and noting that the average reward for the optimal policy, $g^*_s$, is independent of the timesteps spent in the pose (i.e., $g^*_s (T^\pi_s) = g^*_s(T^*_s) \ \forall s$). 

We first focus on the first term in~\eqref{bandit-proof-3}. We know that $\mathbb{E}\left[\mathcal{R}^\mathcal{B}_\pi(T)\right]$ approaches zero if each pose is visited infinitely often in the limit as $T \rightarrow \infty$ provided that $\mathcal{B}$ is a no-regret online learning algorithm:
 \begin{align}
 \label{eq:inf_pose}
     \lim_{T \rightarrow \infty} p_T^\pi (s) > 0, \forall s \in \lbrace 1, 2, \hdots, S\rbrace \quad \Rightarrow \quad \lim_{T \rightarrow \infty} \mathbb{E}\left[\mathcal{R}^\mathcal{B}_\pi(T)\right] = 0
 \end{align}
Thus, it remains to be shown that under $\pi$, all poses are visited infinitely often in the limit as $T \rightarrow \infty$. Note that this statement is equivalent to showing that in the limit as $T \rightarrow \infty$, bandit algorithm $\mathcal{B}$ selects grasps on each pose with non-zero success probability with non-zero probability. Suppose that this was not the case (i.e., that as $T \rightarrow \infty$, $\mathcal{B}$ assigns zero grasp probability to all grasps with non-zero success probability). This would imply that $\mathcal{B}$ only selects grasps with zero success probability, and thus incurs constant regret on its own distribution ($\lim_{T \rightarrow \infty} \mathbb{E}\left[\mathcal{R}^\mathcal{B}_\pi(T)\right] > 0$). This contradicts the initial assumption that $\mathcal{B}$ is a no-regret online learning algorithm, showing that under $\pi$, all poses must be visited infinitely often in the limit as $T \rightarrow \infty$.

Now we shift our attention to the second term in~\eqref{bandit-proof-3}. Given that $\mathcal{B}$ is a no-regret online learning algorithm, it must be the case that $g^\pi_s(T_s^\pi) \xrightarrow[T_s^\pi \to \infty]{} g^*_s (T_s^\pi) \ \forall s$. This implies that in the limit as $T \rightarrow \infty$, $\pi$ and $\pi^*$ have the same success rate on all stable poses. Two policies with the same success rate on all stable poses induce the same Markov chain over $\mathcal{S}$, and thus admit the same stable pose distribution. Thus, $p^\pi_T(s) \xrightarrow[T \to \infty]{} p^*_T(s)$, implying that the second term also approaches 0 as $T \rightarrow \infty$. \qed

\subsection{Proof of Theorem~\ref{theorem:pose-coverage}}
\label{sec:pose-coverage-proof}
Let $X$ denote the random variable that is the number of drops needed until we've reached every stable pose. We assume we have $n$ distinct stable poses. Let $X_k$ be the discrete random variable that represents the number of drops after visiting the $(k-1)$th distinct stable pose to visit the $k$th distinct stable pose. As a base case, $X_1 = 1$ since the first pose we drop into will always be distinct. By linearity of expectation we have
\begin{equation}
    \Ex[X] = \sum_{j=1}^N \Ex[X_j].
\end{equation}

After we visit the $(j-1)$th distinct stable pose we have $N-j+1$ stable poses left that haven't been visited yet. As before, we order the stable poses in order from most likely to least likely: $\lambda_1 > \lambda_2 > \cdots > \lambda_N$. Because the expected number of drops to get into an unseen pose is a geometric random variable, we can upper bound the number of drops needed to visit all stable poses as
\begin{equation}
     \Ex[X] = \sum_{j=1}^N \Ex[X_j] \leq \sum_{j=1}^N \frac{1}{1 - \sum_{i=2}^{j} \lambda_{i-1}}.
\end{equation}
We derive this by first noting that the probability of visiting a new stable pose is $1-\sum_{i \in \text{visited}} \lambda_i$ and the expected number of drops to successfully visit a new stable pose is a geometric random variable. We want to bound the number of drops to visit all stable poses, but we do not know what order we will visit these poses. However, the worst-case outcome (largest number of required drops) will be if we visit them in order from largest drop probability to smallest since this will reduce the probability as quickly as possible which will maximize $E[X]$ since the probabilities are in the denominator. 
The above analysis assumes we are repeatedly able to randomly drop the object, with no accounting for grasp success or failure. Rather than only bounding the number of drops, we want to bound the expected number of grasps to visit all stable poses at least once. 

The grasping strategy will impact how often we need to attempt grasps before we get to drop again. For a worst-case analysis, we assume that there are no topples, $\delta_{s,s'} = 0 \ \forall s, s'$, and assume that each stable pose has only one grasp with non-zero success probability. We will also let $\epsilon$ be the lower bound on the success probability of this grasp across all stable poses. If we just pick grasps uniformly at random on each pose, we have a $1/K$ probability of selecting the best grasp (ie. the one grasp with non-zero success probability) out of $K$ possible grasps. Note that this strategy can only yield strictly lower grasp success rate than \algabbr since any no-regret algorithm (such as UCB or Thompson sampling) will cause \algabbr to pick grasps with non-zero success probability with higher probability than grasps which never succeed. Thus, if we can upper bound the number of grasps under a uniform random grasp strategy this will upper bound the number of grasps for \algabbr. The lower bound on the probability of a successful grasp which leads to a redrop is $\epsilon$. Thus, in the worst-case, we have probability $\epsilon/K$ of getting to redrop and the worst-case expected number of grasps to get a successful grasp is $K/\epsilon$ since this is a geometric random variable. Thus, the expected number of grasps we need to visit all stable poses is no more than
\begin{equation}
    \frac{K}{\epsilon}\sum_{j=1}^N \frac{1}{1 - \sum_{i=2}^{j} \lambda_{i-1}}.
\end{equation}
\qed

\section{Experimental Details}
\label{sec:exp-details}
\textbf{Object Selection:} We choose 7 objects from the set of adversarial objects in Dex-Net 2.0~\cite{mahler2017dex} because these objects had empirically been shown to be difficult to grasp for the Dex-Net policy. Similarly, the recently-introduced EGAD! object dataset~\cite{morrison2020egad} was created to contain objects with few high-quality parallel-jaw grasps. For this dataset, we select all objects for which there exists at least one sampled grasp of quality $\epsilon = 0.1$ on at least one stable pose of the object. Of the 49 objects in the EGAD evaluation dataset, 39 met this criterion.

\textbf{Pose Selection:} For each of the objects, we remove stable poses from the distribution in simulation if they occur with less than a 0.1\% chance or if they do not contain a sampled grasp with quality at least $\epsilon=0.1$. When a pose is removed, the remaining stable pose distribution is renormalized.

\textbf{Grasp Sampling:} We sample a set of $K = 100$ parallel-jaw grasps on the image observation of each pose of each object as in~\cite{mahler2017dex}. This sampling process is done using the depth image grasp sampler from the GQCNN repository and is repeated for up to 10 iterations. At each iteration, the sampled grasps' ground truth qualities are calculated using the robust wrench resistance metric that measures the ability of the grasp to resist gravity~\cite{mahler2018dex}. If no grasps are found with quality of at least $\epsilon = 0.1$, then the sampling process is repeated for another iteration where more grasps are sampled. If a grasp is found with quality at least $\epsilon$, then that grasp is selected along with 99 other grasps chosen at random from the sampled grasps. If the maximum number of iterations are exceeded without finding a grasp with quality $\epsilon$, the stable pose is discarded.

\section{Additional Simulation Experiments}
\label{sec:additional-exps}
\begin{figure*}[htb!]
    \centering
    \includegraphics[width=\linewidth]{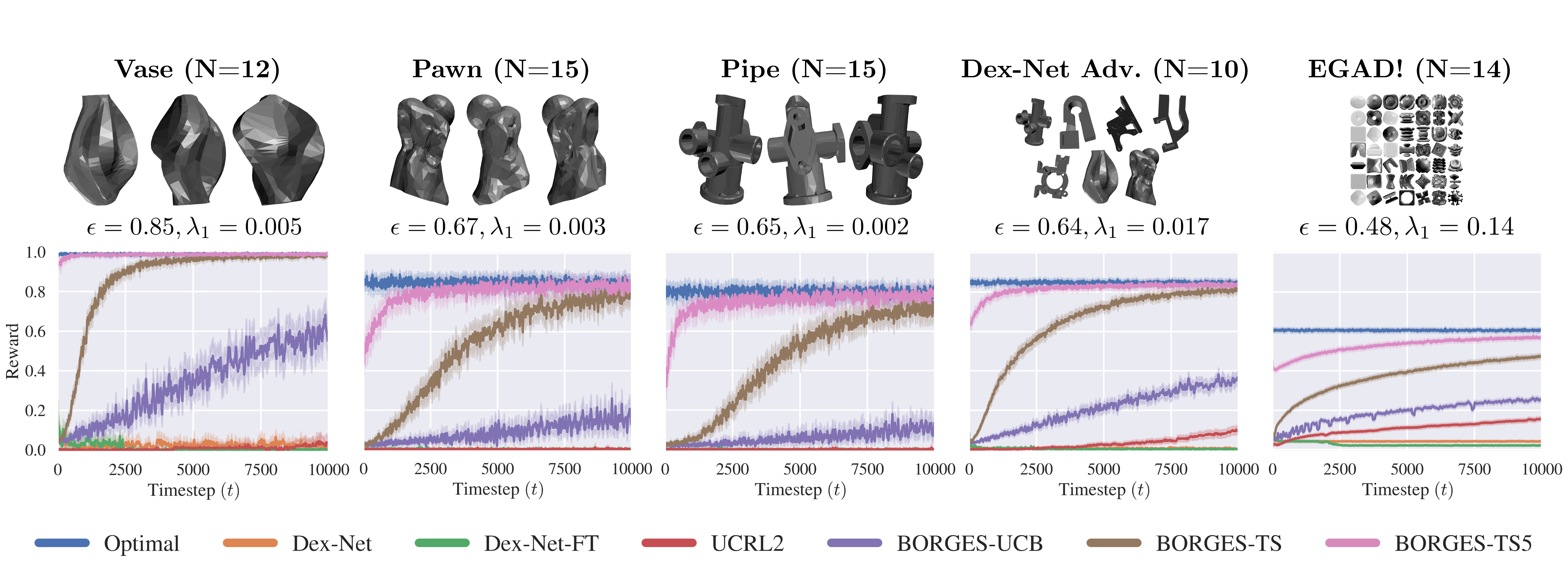}
    \caption{Reward plots for three Dex-Net Adversarial objects as well as aggregated rewards across all Dex-Net Adversarial objects and EGAD! objects in a scenario where no toppling is allowed between poses. Poses without any nonzero quality grasps are discarded. When transitions can only occur due to successful grasps, Exploratory Grasping policies converge to the optimal policy much more quickly than baselines.}
    \label{fig:rewards_no_toppling}
\end{figure*}

We also evaluate \algabbr and baselines in a setting in which toppling is not possible ($\delta_{s, s'} = 0, \ \forall s \neq s$). This exacerbates the difficulty of grasp exploration since an object must be successfully grasped in a given pose for policies to be able to explore grasps in other poses. When evaluating policies, we remove poses that have no grasps with nonzero ground-truth quality and renormalize the stable pose distribution. We emphasize that we perform this step for all policies and do this only to ensure that no poses act as sink states from which the robot can never change the pose in the future (and thus ensure that Assumption~\ref{assum:ergodicity} is satisfied). For the 7 objects from the Dex-Net 2.0 adversarial object dataset, $11\%$ of poses (an average of 1.3 poses per object) were removed. These poses made up an average of $2.8\%$ of the probability mass of the stable pose distribution. These results are shown in Figure~\ref{fig:rewards_no_toppling}. We find that \algabbr again significantly outperforms baseline policies in this setting.

\begin{figure*}[htb!]
    \centering
    \includegraphics[width=\linewidth]{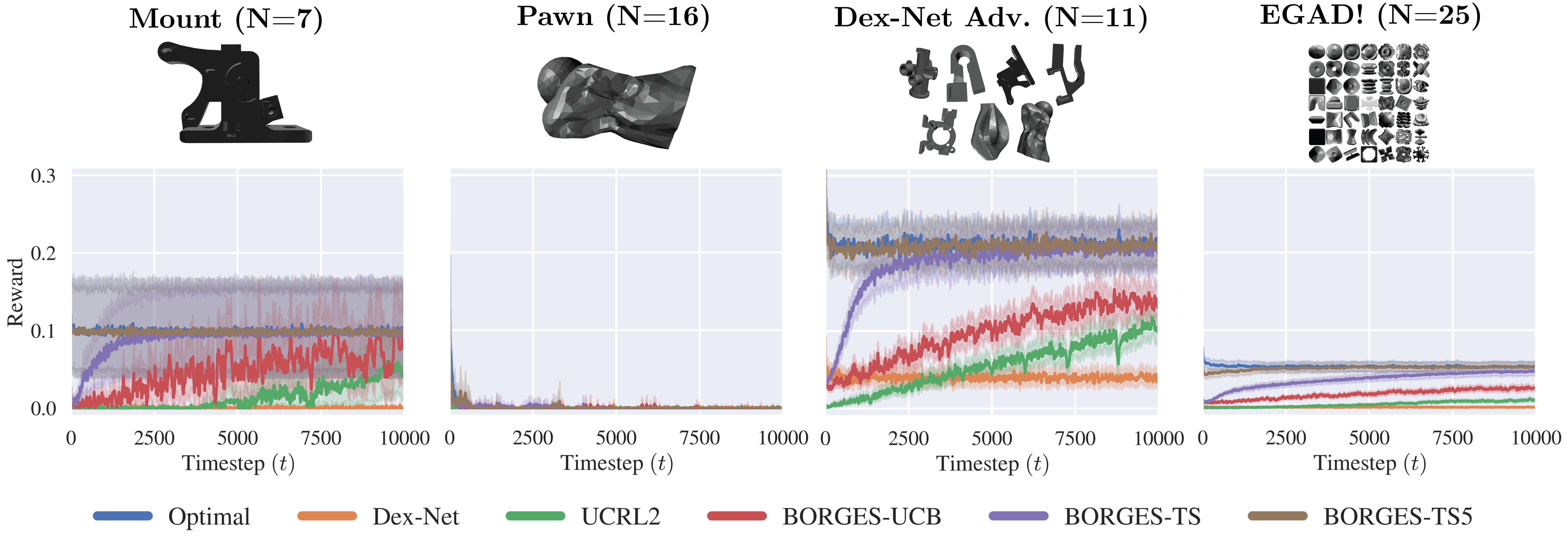}
    \caption{Reward plots for two Dex-Net adversarial objects shown alongside an aggregation of all Dex-Net adversarial objects and EGAD! objects in a scenario where no poses are removed from exploration. In such a case, there is a split between objects without sink states, such as the mount, and objects with sink states, such as the pawn, where no means of transitioning via toppling or grasping is available.}
    \label{fig:all_pose_rewards}
\end{figure*}

We show the importance of Assumption~\ref{assum:ergodicity} even when toppling is included by evaluating policies on all stable poses of the object, regardless of whether a grasp with nonzero ground-truth quality exists or toppling out of each pose is possible. Thus, it is possible that some poses may act as sink states. Figure~\ref{fig:all_pose_rewards} suggests that for objects that do not obey Assumption~\ref{assum:ergodicity}, such as the pawn, all policies find sink states and remain in them. For objects that do obey Assumption~\ref{assum:ergodicity}, but have stable poses that can only transition via toppling, such as the mount, performance decreases for all policies, but the order of convergence is consistent.

\section{Sensitivity Experiments}
\label{sec:sensitivity-exps}
\subsection{Sensitivity to Number of Grasp Samples}
\begin{figure*}[htb!]
    \centering
    \includegraphics[width=\linewidth]{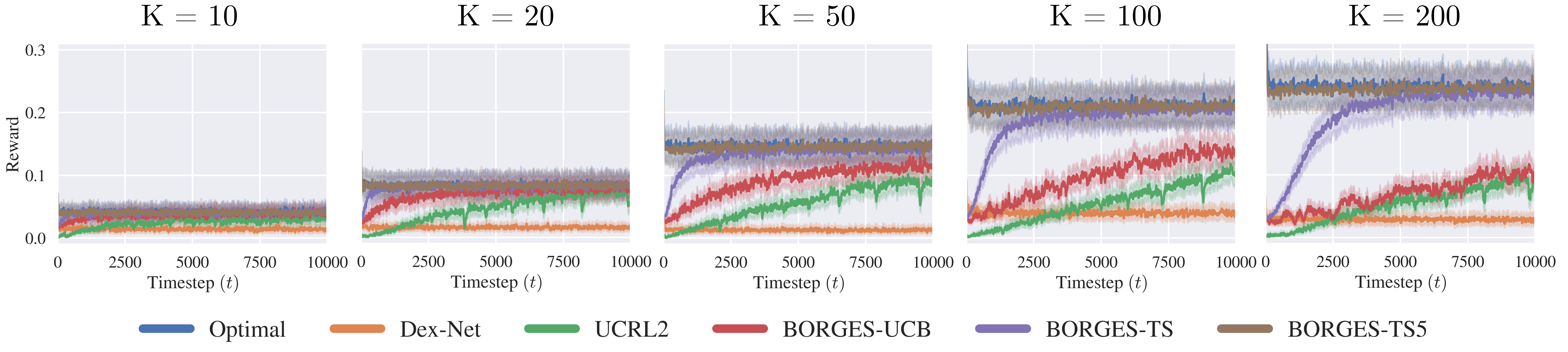}
    \caption{Sensitivity of \algabbr to the number of grasp samples $K$ across the 7 Dex-Net Adversarial objects. With more grasp samples, the likelihood of including a higher-quality grasp increases, but the rate of convergence decreases.}
    \label{fig:k_ablation}
\end{figure*}

We vary $K$, the number of grasp samples per object pose, within the range $K = \lbrace 10, 20, 50, 100, 200 \rbrace$ across the 7 objects in the Dex-Net Adversarial object set. The results in Figure~\ref{fig:k_ablation} suggest that as $K$ increases, the likelihood of sampling a high-quality grasp on each pose also increases. With more grasps to consider, each policy takes longer to converge to the optimal policy, but the order of convergence between policies is consistent.

\subsection{Sensitivity to \probname Parameters}
\begin{figure*}[htb!]
    \centering
    \includegraphics[width=\linewidth]{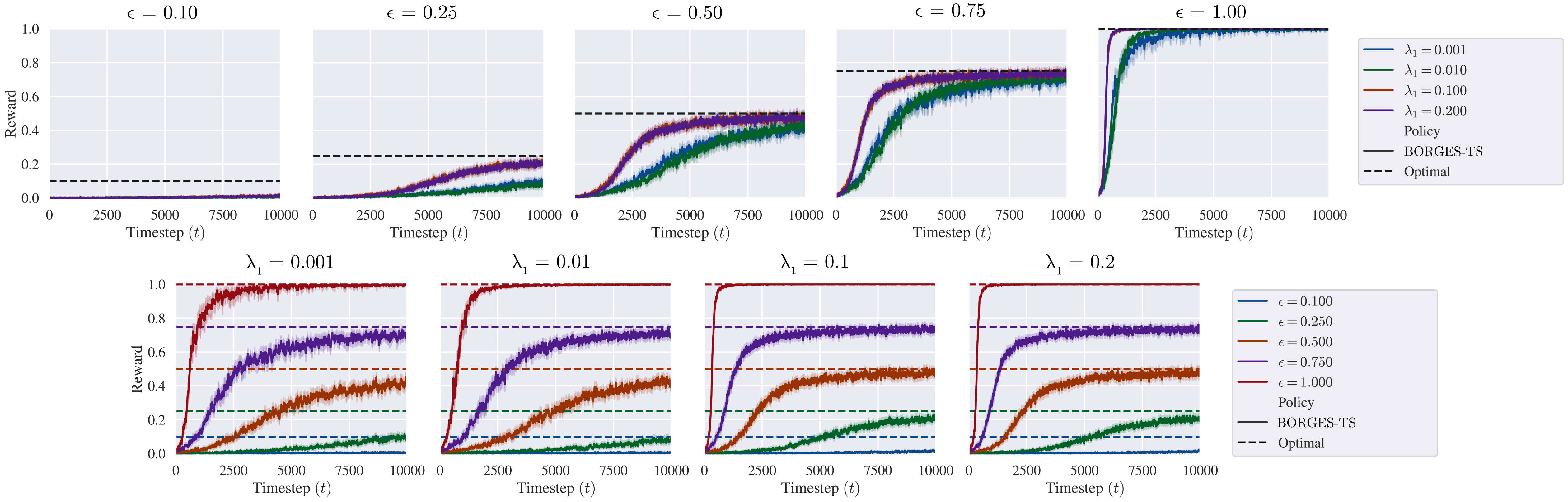}
    \caption{Sensitivity of \algabbr to the \probname parameters $\epsilon$ and $\lambda_1$, as shown across 20 synthetic objects. Unless $\epsilon$ or $\lambda$ is low, \algabbr quickly converges to the optimal policy. However, when either is low, particularly $\epsilon$, the policy converges much more slowly, taking even more than the 10,000 timesteps shown for some combinations.}
    \label{fig:sensitivity}
\end{figure*}

We perform sensitivity analysis of \algabbr to the \probname parameters $\epsilon$ and $\lambda_1$. For these experiments, we evaluate the policy using a set of synthetic objects with $\lambda_1 = \lbrace 0.001, 0.01, 0.1, 0.2 \rbrace$ and $\epsilon = \lbrace 0.1, 0.25, 0.5, 0.75, 1.0 \rbrace$ and for simplicity consider the case in which toppling is not possible. In each case, we choose a single grasp on each pose to have reliability $\epsilon$ with all other grasps having a mean parameter of $0$. The results are shown in Figure~\ref{fig:sensitivity}. These results suggest that unless $\epsilon$ or $\lambda_1$ are low, \algabbr quickly converges to the optimal policy. In particular, $\epsilon$ has an outsized effect on the accumulated reward; with $\epsilon \leq 0.10$, we observe that the policy fails to approach the optimal policy regardless of $\lambda_1$ through the first 10,000 timesteps.

\section{Initial Physical Experiments Details}
\label{sec:phys-exps-supp}
In physical experiments, we evaluate \algabbr on two challenging 3D printed objects and compare performance with the Dex-Net policy from Section~\ref{sec:sim-exps}. The object is placed in front of an ABB Yumi robot with a parallel-jaw gripper and overhead Photoneo PhoXi S camera. At each timestep, we capture a depth image observation from the camera and estimate if the current object stable pose has been seen previously with the following stable pose change detection procedure. Specifically, we cache the first depth image seen for each new pose, and compare the current depth image at each timestep to the cached depth images. We segment out the background and project the depth image into a point cloud representation. We then mean-center the current point cloud, apply $7200$ rotations around the z-axis, and measure the chamfer distance between each resulting rotated point cloud and the point cloud representations of each of the cached depth images. We classify each point in the current point cloud as an inlier if the closest point in a previous point cloud is less than $0.02$ mm away. If at least $80\%$ of the points are inliers, we classify the two point clouds as belonging to the same stable pose. If the pose has not been seen previously, we add the depth image to the cache, sample a set of grasps, and calculate the Dex-Net predicted quality values for each pose. Otherwise, we recall the previously sampled grasps for the given pose and transform them by the relative transform between the cached and current images. A grasp is executed according to the Dex-Net or \algabbr policy; if the grasp is successful, the object is lifted, rotated by 30 degrees around an axis sampled uniformly at random from 3D unit vectors, and released. If the grasp is unsuccessful, the object may topple into another pose or may remain in the same pose. This process repeats for 200 timesteps per object. After each grasp attempt, we move the object to the center of the workspace to prevent errors in the stable pose change detection procedure, but strongly suspect that this procedure will likely be unnecessary with a more constrained workspace or a more robust stable pose change detection procedure, which we will implement in future work.

\begin{figure}
    \centering
    \includegraphics[width=\linewidth]{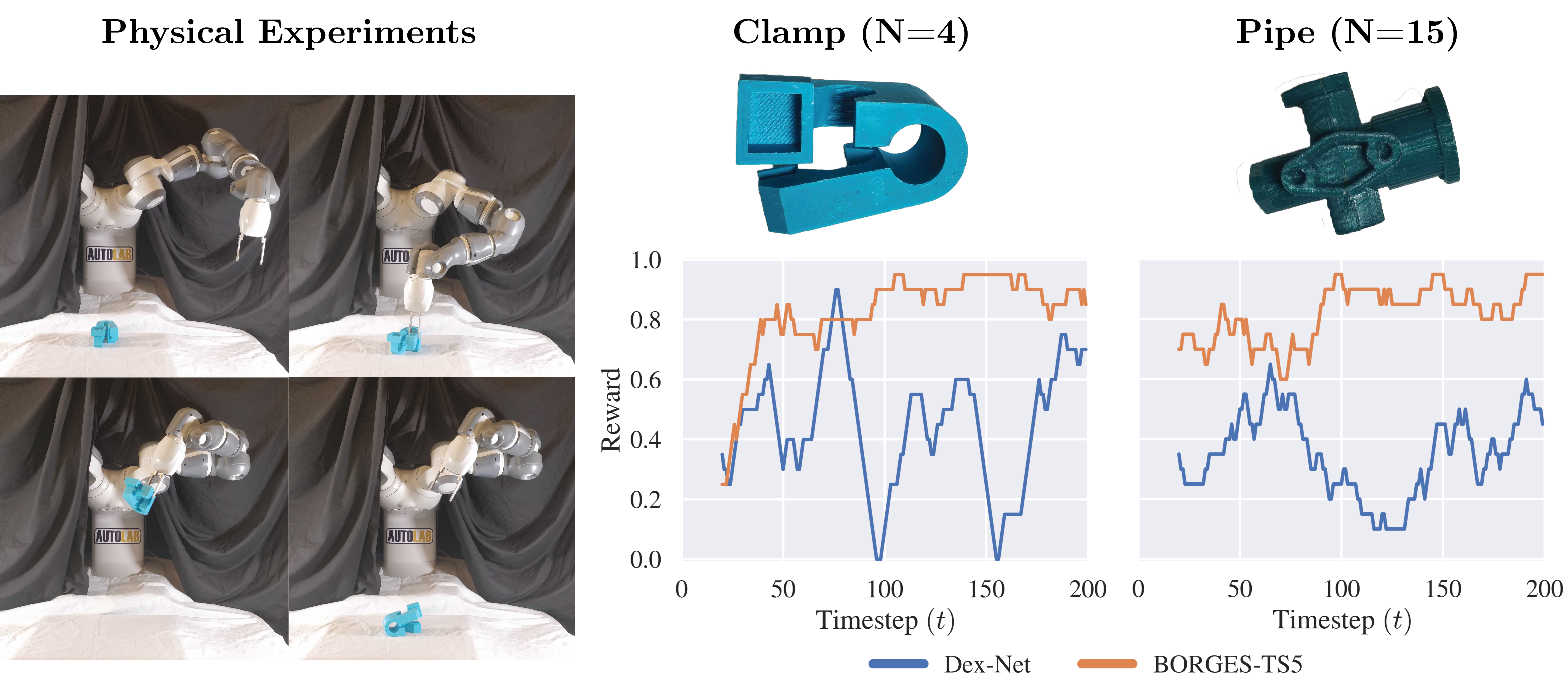}
    \caption{Experiment setup and learning curves for the Dex-Net and \algabbr-TS5 policies for the clamp and pipe objects across 200 grasp attempts (smoothed with a running average of 20 attempts). The robot attempts to grasp the object at each timestep and, if it succeeds, rotates and drops the object to sample from the stable pose distribution (left). \algabbr-TS5 quickly converges within 100 attempts on both objects, indicating that it finds grasps that succeed nearly every time for each pose. Dex-Net's performance remains uneven, indicating that it finds high-quality grasps for some poses, but not others.}
    \label{fig:physical_rollouts-supp}
\end{figure}

In Figure~\ref{fig:physical_rollouts-supp} we show a larger version of Figure~\ref{fig:physical_rollouts} from Section~\ref{sec:phys-exps}. As described in Section~\ref{sec:phys-exps}, we find that \algabbr is able to significantly increase the grasp success rate of a Dex-Net policy, achieving success rates of $0.89$ and $0.87$ on the clamp and pipe, respectively, as compared to $0.49$ and $0.37$ for the Dex-Net policy with just 200 grasp attempts in the real world. This effect highlights the importance of a policy that is able to learn online from successful and failed grasp attempts; Dex-Net does not learn online so continuing to attempt grasps will lead to the same high-variance behavior over time, but \algabbr continues to stabilize as it approaches optimal performance across all poses.

\end{document}